\documentclass[10pt,twocolumn,letterpaper]{article}

\usepackage[pagenumbers]{cvpr} 

\definecolor{cvprblue}{rgb}{0.21,0.49,0.74}
\usepackage[pagebackref,breaklinks,colorlinks,allcolors=cvprblue]{hyperref}

\def\paperID{*****} 
\def\confName{CVPR}
\def\confYear{2026}

\newcommand{\name}{\mbox{ExpressEdit}} 
\usepackage{cuted} 

\usepackage{tcolorbox}
\usepackage{xcolor}

\definecolor{PhotoshopUIGray}{HTML}{535353}   

\newcommand{\photoshop}[1]{\textcolor[HTML]{ff0080}{#1}}
\newcommand{\plugin}[1]{\textcolor[HTML]{3571de}{#1}}

\newtcolorbox{promptbox}[1][]{
  colback=gray!5,        
  colframe=PhotoshopUIGray,     
  coltitle=white,        
  fonttitle=\bfseries,   
  boxrule=0.5mm,         
  arc=2mm,               
  title=#1,              
  fontupper=\mdseries\small\setlength{\parskip}{0.2em}, 
  #1                     
}

\usepackage{CJK}
\newcommand{\chinese}[1]{\begin{CJK}{UTF8}{gbsn}#1\end{CJK}}
\newcommand{\japanese}[1]{\begin{CJK}{UTF8}{min}#1\end{CJK}}

\title{ExpressEdit: Fast Editing of Stylized Facial Expressions with Diffusion Models in Photoshop}

\author{Kenan Tang, Jiasheng Guo, Jeffrey Lin, Yao Qin\\
University of California, Santa Barbara\\
{\tt\small kenantang@ucsb.edu, yaoqin@ucsb.edu}
}

\begin{document}
\maketitle

\begin{strip}
\centering
    {\includegraphics[width=1\textwidth]{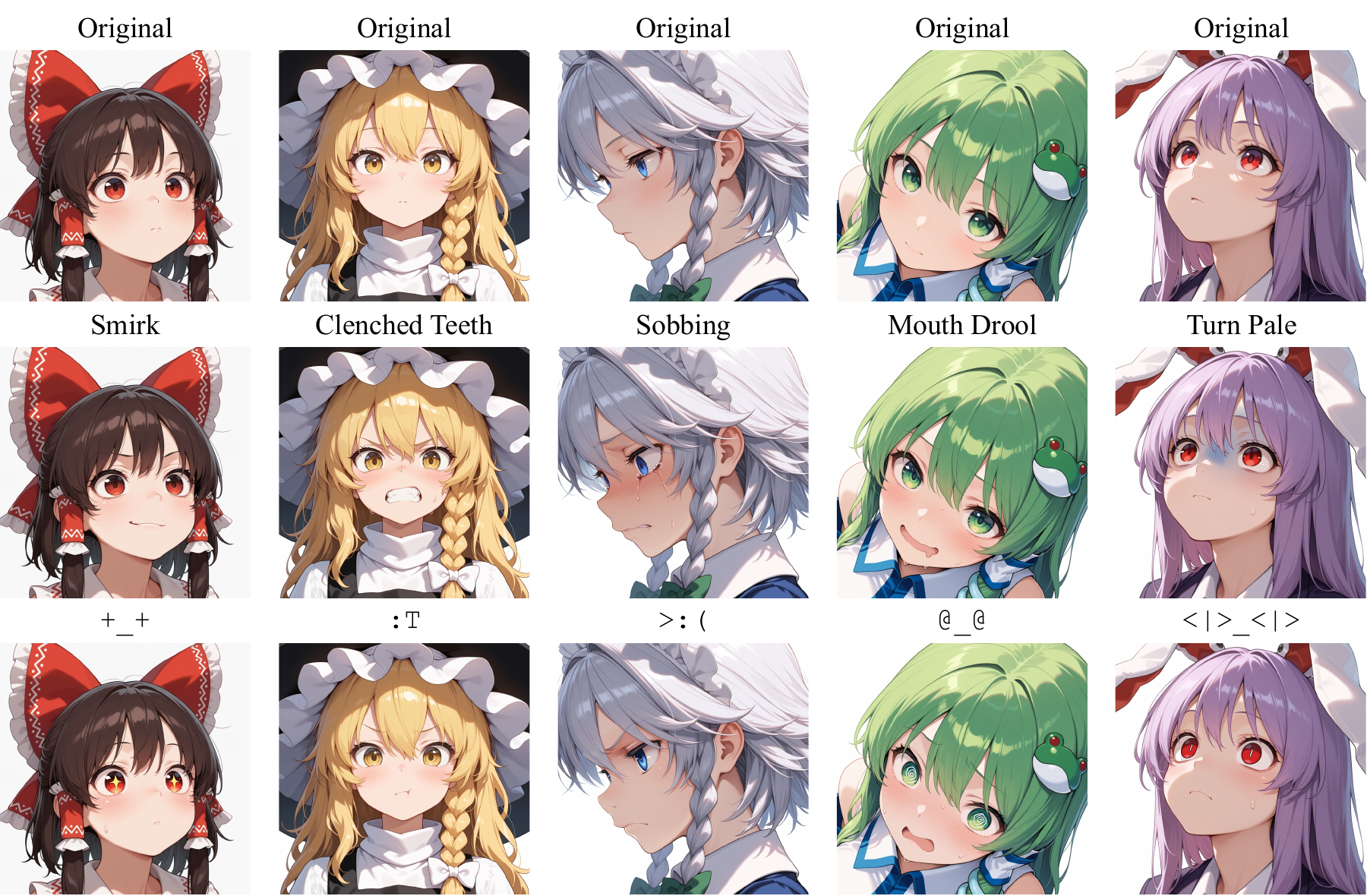}}
\captionof{figure}{\textbf{\name\ can generate diverse, stylized expressions on an original image.} The first row shows the original image, and the second and third rows show the edited images, with the user-specified expression above each image. \name\ can handle both detailed multi-word descriptions (such as ``clenched teeth'') and emoticons (such as ``\texttt{@\_@}''), generating stylized depictions of expressions with ease.}
\label{fig:diverse}
\end{strip}

\begin{abstract}
    Facial expressions of characters are a vital component of visual storytelling. While current AI image editing models hold promise for assisting artists in the task of stylized expression editing, these models introduce global noise and pixel drift into the edited image, preventing the integration of these models into professional image editing software and workflows. To bridge this gap, we introduce ExpressEdit, a fully open-source Photoshop plugin that is free from common artifacts of proprietary image editing models and robustly synergizes with native Photoshop operations such as Liquify. \name\ seamlessly edits an expression within 3 seconds on a single consumer-grade GPU, significantly faster than popular proprietary models. Moreover, to support the generation of diverse expressions according to different narrative needs, we compile a comprehensive expression database of 135 expression tags enriched with example stories and images designed for retrieval-augmented generation. We open source the code and dataset to facilitate future research and artistic exploration.\footnote{\url{https://github.com/kenantang/ExpressEdit}}
\end{abstract}

\section{Introduction}

Facial expressions on characters are vital for visual storytelling~\cite{su2007personality, lasseter1998principles, porter2000site, zhang2021influence}, yet creating detailed expressions is a time-consuming process, even with the assistance of professional software~\cite{lau2009face, abdrashitov2020interactive, chen2020deep, amini2013hapfacs}. Besides realistic human faces, visual storytelling commonly uses 2D or 3D animation characters, which necessitates the generation and editing of stylized expressions on these characters~\cite{aneja2016modeling, aneja2018learning}. 

AI tools are increasingly applied to visual content generation~\cite{Atzmon_2025_BMVC, Baranwal_2025_ICCV, Tang_2025_ICCV, cardoso2024re} and storytelling~\cite{Shin_2025_ICCV, Vitasovic_2025_ICCV, ghorbani2025aether, akdemir2025plot, shin2024lost}. Many tools can already assist artists on generating or editing realistic expressions~\cite{zou20244d, zhang2024emotalker, thambiraja20233diface}. Despite technical improvements, stylized expression editing remains challenging for two reasons. One reason is that as most expression editing systems are tailored to realistic expressions, the depiction of stylized expressions are sometimes interfered by real face features, resulting in artifacts that are neither realistic nor stylistic~\cite{jiang2026emojidiff}. Another reason is the failure to precisely control proportions and positioning of facial features, like the eye-to-mouth distance~\cite{kanade2000comprehensive}. This precision is essential for conveying a character's identity~\cite{zhang2021influence, allen2021nasal, chen2025exploring}, such as their age~\cite{glocker2009baby} or personality~\cite{carter2016designing}, but even the latest proprietary models fail to follow instructions with precise numerical values faithfully (\Cref{fig:half_iris}).

Latest image editing models, such Nano Banana 2~\cite{google2026nanobanana2}, are increasingly good at generating both realistic and stylistic images~\cite{arena2026imageedit}, mitigating the two expression-specific challenges above to a certain extent. However, from a practitioners' viewpoint, we identify three more persisting weaknesses that still cause significant inconvenience for users:

First, these models primarily rely on textual prompts for image generation. With this restriction, users have to come up with detailed descriptions of expressions~\cite{wang2024promptcharm, liu2022design}, otherwise the generated results lack diversity~\cite{wadinambiarachchi2024effects}. This requirement on the prompt quality poses a cognitive burden for users and slows down the creation process~\cite{tang2024s, huang2025promptnavi, chong2025prompting}.

Secondly, these models suffer from noise artifacts or watermarks~\cite{gowal2025synthid, googledeepmind2026synthid}. These artifacts are visually disturbing, and the noise can be amplified in consecutive edits, consistently degrading the image quality (\Cref{fig:eight_steps}). 

Thirdly, while some models have been integrated into professional editing software like Photoshop~\cite{adobe2026generativefill}, these models cause undesired resolution changes and pixel drifts, worsening user experience (\Cref{fig:local}). This cumbersome integration prevents the user from enjoying the benefits of AI integration into professional software.

To address these weaknesses, we propose \name, a fully open-source Photoshop plugin that edits diverse expressions cleanly and seamlessly (\Cref{fig:diverse}). Helped by numerous native Photoshop operations, the user gains precise control over the size and location of facial elements. Furthermore, \name\ is equipped with an expression database of 135 expression tags, supporting retrieval-augmented generation (RAG) that lowers the entry barrier to new users without prior knowledge of its tag-based prompt format. Despite its high output quality, \name\ edits each expression within 3 seconds on a single consumer-grade GPU, a latency far lower than that of all proprietary models we examined. With all these advantages, \name\ provides smooth expression editing experience for both beginners and professionals alike.

In the sections below, we elaborate on the design of \name\ (\Cref{sec:the-expressedit-plugin}) and demonstrate its advantages over latest representative proprietary models (\Cref{sec:advantages-of-expressedit}).

\section{The ExpressEdit Plugin}
\label{sec:the-expressedit-plugin}

\begin{figure*}
  \centering
  \begin{minipage}{0.63\textwidth}
      \begin{subfigure}{1\linewidth}
          \centering
          \includegraphics[width=1\textwidth]{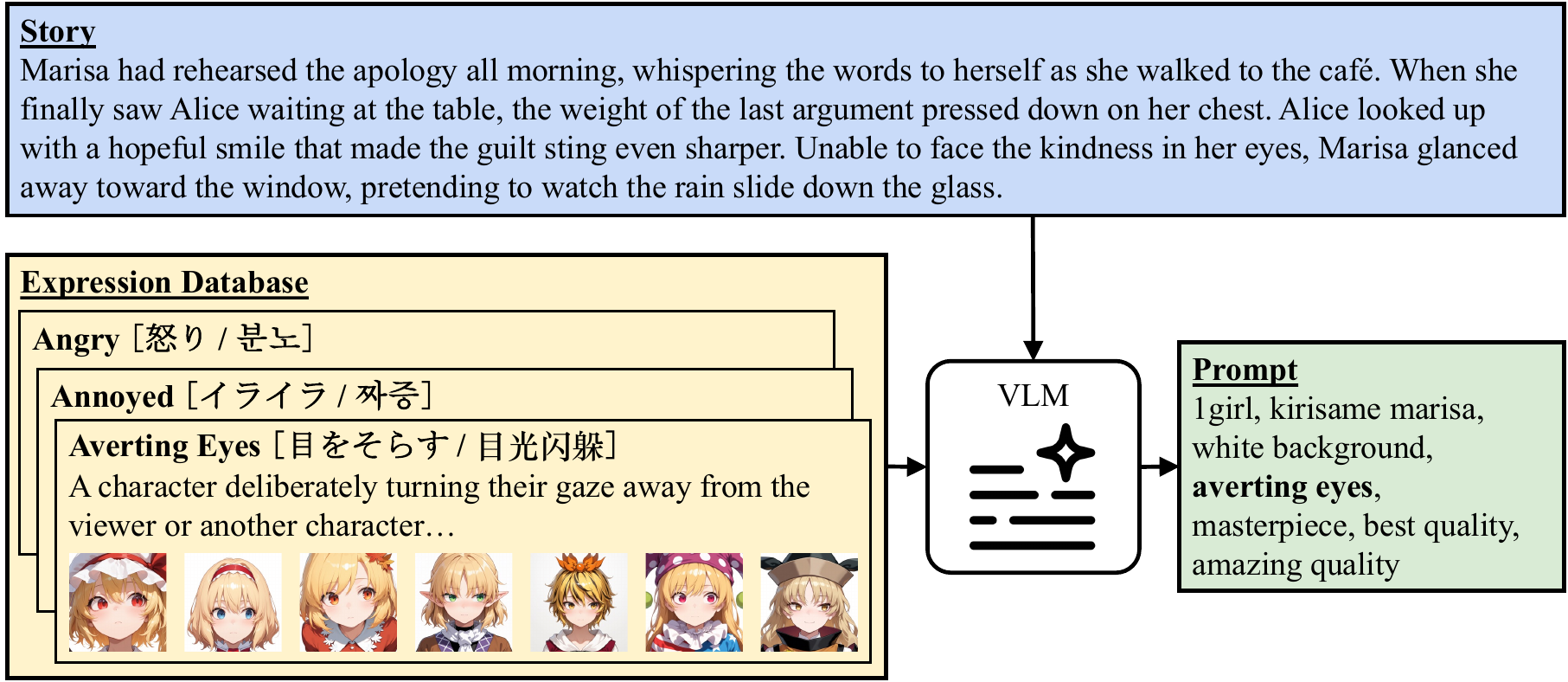} 
          \caption{The retrieval-augmented prompt generator. Only the expression tag (bold) is generated by the VLM. The template words before and after the expression tag are pre-specified.}
          \label{fig:pipeline-text}
      \end{subfigure}
      \begin{subfigure}{1\linewidth}
          \centering
          \includegraphics[width=1\textwidth]{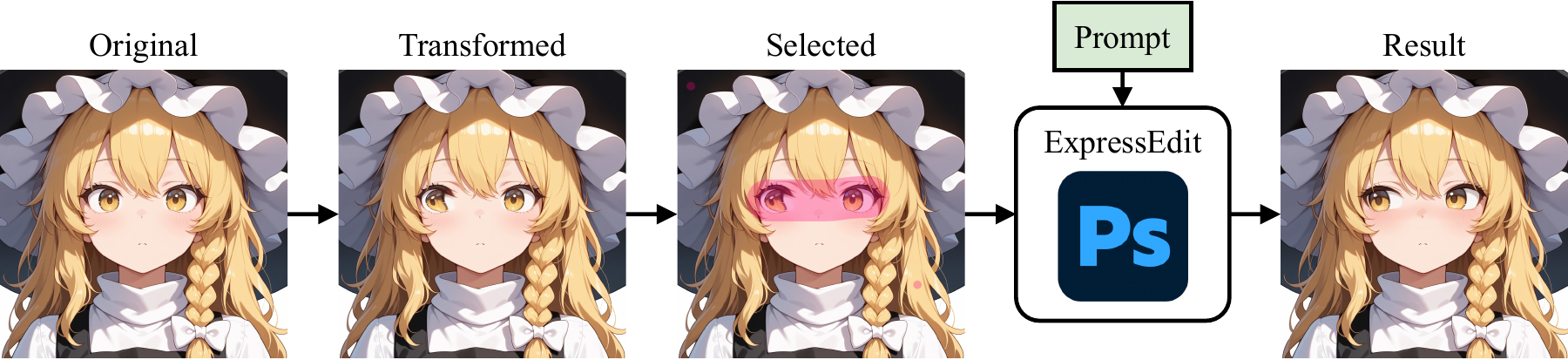} 
          \caption{The expression editor, enhanced by the Liquify transformation in Photoshop in this example.}
          \label{fig:pipeline-image}
      \end{subfigure}
  \end{minipage}
  \hfill
  \begin{minipage}{0.35\textwidth}
      \begin{subfigure}{1\linewidth}
          \centering
          \includegraphics[width=1\textwidth]{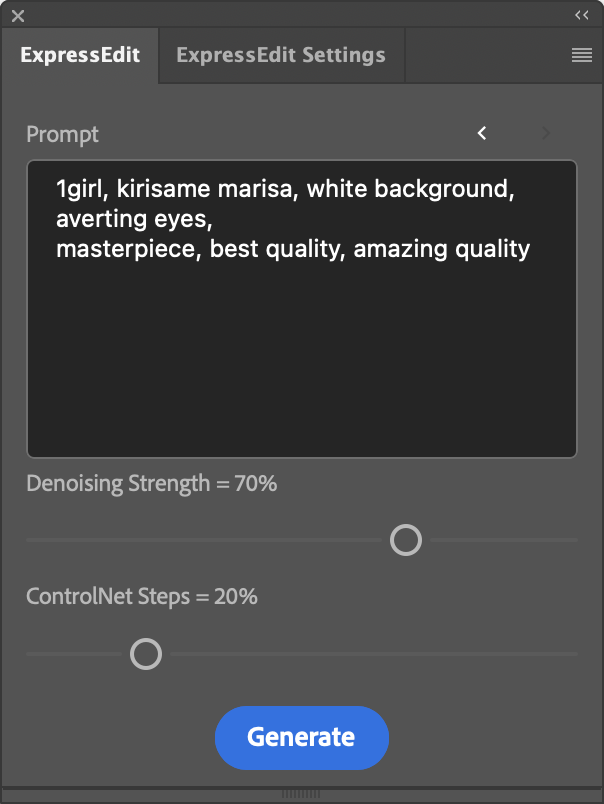} 
          \caption{The streamlined \name\ panel in Photoshop.}
          \label{fig:panel}
      \end{subfigure}
  \end{minipage}
  
  \caption{\textbf{\name\ consists of two consecutive pipelines for a user-friendly yet professional editing experience.} The prompt generation pipeline takes in a story paragraph and uses a VLM to retrieve relevant expression tags from a multi-modal expression database we curate. The relevant expression tags are inserted into a prompt, which is used in the image editing pipeline. The image editing pipeline starts by the user applying coarse transformations (such as Liquify) and casual selections on the original image, taking at most a few seconds of manual effort. Then, combined with the prompt, \name\ robustly generates high-quality expressions based on the inputs.}
  \label{fig:pipeline}
\end{figure*}

\Cref{fig:pipeline} visualizes two components of the \name\ Photoshop plugin, which are the retrieval-augmented prompt generator and the expression editor. The prompt generator converts the user intent, a story in this case, into an expression tag (``averting eyes''), which is inserted into a customizable prompt template, consisting of a prefix describing the image content and a suffix controlling the image style (\Cref{fig:pipeline-text}). Then, the user provides the prompt and an image input into the expression editor, with optional transformations and a required selection on the edited region (\Cref{fig:pipeline-image}). Next, the user clicks ``Generate'' on the frontend panel (\Cref{fig:panel}). Finally, a diffusion-model-based backend will edit the image and return the result as a new image layer.

In the subsections below, we explain the design and usage of each component in detail.

\subsection{Retrieval-Augmented Prompt Generator}
\label{sec:retrieval-augmented-prompt-generator}

The expression editor component of \name\ requires a tag-based format of prompts for its diffusion model backend. Since the format differs from natural language description, it poses a learning barrier for new users. To lower this learning barrier, we draw inspiration from existing tools in the community~\cite{mirabarukaso2026characterselect} and design a retrieval-augmented generation (RAG) system, which allows users to easily retrieve the tags. A RAG system bridges the gap between large and small text generation models~\cite{li2025okbench}, providing additional convenience for users with different levels of compute resource. 

We constructed an expression tag database for the RAG system. The database consists of the following 6 parts:

\paragraph{Expression Tags.} We obtained the expression tags from the Danbooru official website of face tags and eye tags~\cite{danbooru2026facetags, danbooru2026eyestags}. Danbooru tags are the basis of the prompt of the text formats for training Illustrious~\cite{park2024illustrious}, the base model of the fine-tuned image generation model in the backend. We manually choose the tags that can assist expression generation, discarding less informative tags such as ``blue eyes'' or ``eye patch.'' Then, tags that can potentially be used to generate explicit content are manually filtered. In this step, we obtain 135 expression tags.

\paragraph{Example Images.} Example images were automatically generated based on 5 original images of different characters (\Cref{fig:diverse}). In this automatic process, no Photoshop transformations (\Cref{sec:expression-editor}) were applied, and the selection was a fixed-size circle that covered the face of the character on each original image. For each expression tag, we repeat the generation 5 times with different random seeds, resulting in 3,375 total edited images.

\paragraph{Transformation-Free-Editing Flag.} For users to identify which expressions can be edited without Photoshop transformations and speed-up the editing of such expressions by optionally skipping transformations, we inspected the example images that are generated without any Photoshop transformations. We found that 35 out of 135 expression tags cannot be reliably edited without Photoshop transformations. One example is ``averting eyes'', where the irises of the characters cannot be moved in arbitrary directions and magnitude with transformation-free editing. While we later show how these expressions can be robustly handled by quick transformations (Sections \ref{sec:responsive-and-precise-edits} and \ref{sec:quick-synergy-with-the-liquify-tool}), we flag these expressions as unable to be edited in a transformation-free manner. For these expressions, we used \name\ to manually create a smaller set of references, which we put into the plugin documentation instead of the database. Examples of 7 different characters are shown in \Cref{fig:pipeline-image}.

\paragraph{Definition.} The definition for each expression tag was obtained from the official Danbooru website. The definition explains the expression tag and specifies which images should or should not be tagged for an expression (\Cref{fig:story_prompt_and_example}).

\paragraph{Alternative Tags.} Danbooru also provides alternative tags for each tag. These alternative tags are Pixiv tags~\cite{pixiv2026encyclopedia} for each expression, or a simple translation of the expression tag into Chinese, Japanese, or Korean. Since Pixiv is also a popular website among artists, we obtained these alternative tags from the Danbooru website and incorporated them into the dataset. Some expression tags do not have official Danbooru alternative tags, so we manually examined the Pixiv Encyclopedia~\cite{pixiv2026encyclopedia} to find appropriate candidates. In the limited cases where candidates are not found, we translate the expression tags, with strict adherence to the format of existing tags. A total of 332 alternative tags were obtained in this process.

\paragraph{Example Stories.} To inspire users and facilitate the retrieval process, we also generated 5 example stories for each tag with Gemini 3 Flash~\cite{google2025gemini3flash}. The process is repeated for Chinese, English, Japanese, and Korean. The language choice aligns with the existing languages on the Danbooru website, and more language can be easily included. The process results in 2,700 short stories. The generation prompt and example stories are shown in \Cref{fig:story_prompt_and_example}. To best invoke the creativity of LLMs, we generate multiple stories in parallel in a single dialog turn~\cite{tang-etal-2024-creative}.

\begin{figure}[h]
\centering
\begin{promptbox}[title=Story Generation Prompt]
Expression Tag: \japanese{+\_+}

Definition:

When a character's eyes light up in excitement, usually with a yellow four-pointed sparkle in the center. Sparkles may also appear around their head.

Do not confuse with star-shaped pupils, which is for literal four/five-pointed star shaped eyes, or for characters who have naturally cross-shaped pupils, such as Nia Teppelin or Chinchou.

Alternative Tags: [``\japanese{目がしいたけ}'', ``\japanese{しいたけ目}'', ``\chinese{星星眼}'', ``\chinese{两眼放光}'']

Provide a short story background (3 to 5 sentences) that make a character do this expression. Generate natural stories, without explicitly referring to the expression. Write the story in English, and repeat 5 times. Only output 5 stories, each starting with a number from 1 to 5, followed by a period.
\end{promptbox}
\begin{promptbox}[title=Generated Stories]
1. As the master chef lifted the silver lid, the aroma of the legendary golden truffle pasta wafted through the dining hall. Young Elara had waited three years for this reservation, having saved every coin from her apprenticeship to afford a single plate. When she saw the perfectly glazed noodles shimmering under the chandelier, her breath caught in her throat.

2. Leo stood at the edge of the hanger as the tarp was ... 
\end{promptbox}
\caption{\textbf{Creative short stories are included in \name\ to assist retrieval-augmented generation (RAG).} The prompt template and an example story are shown here.}\label{fig:story_prompt_and_example}
\end{figure}

We have a rich database of expression tags, far exceeding the number of common categorization systems~\cite{aneja2016modeling, aneja2018learning}. We also include emoticons~\cite{park2013emoticon, derks2008emoticons, aldunate2017integrated}, which vividly correspond to stylized expressions. Well-grounded in existing Danbooru and Pixiv tags, these tags should be familiar to experience practitioners of digital painting.

Given the database, a user can conveniently use a VLM to retrieve the tag that is relevant to their stories, ideas, or specific editing instructions. The database can be provided as an input to the VLM using various context engineering techniques~\cite{mei2025survey}. By converting free-form user intent into structured expression tags, \name\ refines the prompt into a format suitable for the image generation model, mitigating the prompt sensitivity~\cite{mo2024dynamic, hua-etal-2025-flaw} of multi-modal generative models that potentially degrades image quality.

\subsection{Expression Editor}
\label{sec:expression-editor}

While AI-based image generation models are usually integrated into Gradio~\cite{gradio2026imageeditor, automatic11112026stable, lllyasviel2026forge} or ComfyUI~\cite{comfyui2026maskeditor} interfaces, the simplistic brush and mask functionalities in these interfaces are inconvenient for fine-grained expression control. Hence, we created a Photoshop plugin using the Adobe UXP Developer Tool~\cite{adobe2022uxpdevtool}, with SPICE~\cite{tang2025spice} as the diffusion-model-based image editing backend. The small number of hyperparameters in SPICE enables a lightweight and more direct integration into Photoshop compared to other contemporary methods~\cite{liu2025magicquill, liu2025magicquillv2}. In the rest of the paper, we use \photoshop{magenta} and \plugin{blue} to distinguish between \photoshop{native Photoshop operations} and \plugin{backend-related operations} for clarity. References for official Photoshop documentations from Adobe are provided after the first mention of each Photoshop operation. For users already familiar with professional editing software, learning backend operations will cost little time.

After obtaining the prompt, the user needs to take the following steps to edit the expression. First, the user needs to change the \plugin{prompt} in the plugin prompt box (\Cref{fig:panel}). Then, the user can apply Photoshop transformations to the input image that roughly changes the expression as a hint to the edited outcome. For example, the \photoshop{Liquify}~\cite{adobe2026liquify} transformation can be used to move the right iris to the right. This step can be skipped if the expression has been flagged as transformation-free (\Cref{sec:retrieval-augmented-prompt-generator}), allowing beginners to achieve quality results with only straight-forward selection. Next, the user needs to apply a \photoshop{selection}~\cite{adobe2023getstartedselections} (shown as a transparent magenta color patch in the Photoshop interface) to cover the region to be edited. When only the eyes or the mouth is relevant to the expression, the selection should only cover the relevant region, with optional \plugin{context dots} recommended by SPICE~\cite{tang2025spice}. Finally, the user can click \plugin{Generate} and directly merge the generated new layer onto the original image by \photoshop{Merge Visible}~\cite{adobe2026mergelayers}.

We implemented the two major hyperparameters (\plugin{Denoising Strength} and \plugin{ControlNet Steps}) from SPICE, in order to support advanced editing scenarios. However, keeping these two hyperparameters at their default values (shown on the panel) leads to robust results. Other hyperparameters, such as \plugin{sampling steps} (\Cref{sec:fast-infererence-with-speed-up-loras}), can be adjusted in the \plugin{\name\ Settings} panel. For instructions on how these parameters can be used, we refer interested readers to the original SPICE paper~\cite{tang2025spice}.

The \name\ plugin was implemented in Version 27.4.0 of Photoshop~\cite{adobe2026photoshopreleasenotes}. While the current version of \name\ only supports Photoshop, both the frontend and the backend code is open source, and the plugin can be migrated to free image editing software such as Krita~\cite{krita2026pythonscripting}. For the SPICE backend, we use WAI-illustrious-SDXL as the base model~\cite{wai07312025waiillustrious} and a midsize Canny edge ControlNet model of SDXL as the ControlNet model~\cite{lllyasviel2023sdcontrolcollection}.

\subsection{Baseline Models}

We chose FLUX.2 [max]~\cite{bfl2025flux2}, GPT~\cite{openai2025chatgptimages}, Grok~\cite{xai2026grokimagine}, Nano Banana 2 Fast (without reasoning), and Nano Banana 2 Pro (with reasoning)~\cite{google2026nanobanana2} as baseline models. These models provide convenient image editing functionality via text prompts, easily accessible on their respective web interfaces. This selection covered highly ranked and popular models on the Image Edit Arena~\cite{arena2026imageedit}.

We exclude recent open-source, local models such as Qwen Image Edit~\cite{qwen2025imageedit2511, wu2025qwen} and FLUX.2 [dev]~\cite{bfl2026flux2dev}, because these models are forbiddingly hard to use for practitioners without high-end compute resources. The full version of either model without quantization requires more than 50 GB VRAM to run, and the inference time is over 2 minutes with the recommended number of inference steps, tested on our 3 NVIDIA RTX A6000 GPUs. As a comparison, using a single consumer-grade NVIDIA GeForce RTX 4090 GPU with 24 GB of VRAM, the full version of ExpressEdit completes inference within 5 seconds, and the latency can be further reduced to below 3 seconds with a speed-up LoRA (\Cref{sec:fast-infererence-with-speed-up-loras}).

Due to the page limit, we cannot visualize exhaustively the tests we performed on these models. Since our method is completely free and open source, and all baseline methods are easily accessible, we encourage the readers to independently verify that the presented results are not cherry-picked, and that our qualitative observations align with general user experience.

\section{Advantages of \name}
\label{sec:advantages-of-expressedit}

In the subsections below, we discuss various advantages of \name\ over baseline models.

\subsection{Succinct but Informative Expression Tags}

Our preliminary experiments with various VLMs showed that the expression database we constructed could help condensing long user intents into succinct expression tags. To the best of our knowledge, there is not a dataset that maps user intents to ground-truth expression tags. Hence, we did not conduct a quantitative evaluation on the text pipeline. After all, once users have inevitably gotten familiar with the expression tags, they can directly start from the expression editor by manually providing tags, without relying on the retrieval-augmented prompt generator.

In the following subsections, we show how \name\ delivers superior results with user-provided expression tags. Note that expression tags could be combined (\Cref{fig:large}) for even richer expressions. Moreover, the eyes and mouth can be individually edited to create more expression combinations (\Cref{fig:liquify}). However, to reduce confounding factors, most experiments below are conducted with the single expression tag ``smile'' for \name\ and the prompt ``Make her smile'' for baseline models.

\subsection{Clean Edits without Degradation}
\label{sec:clean-edits}

\begin{figure}
  \centering
  \begin{subfigure}{1\columnwidth}
      \centering
      \includegraphics[width=0.796\columnwidth]{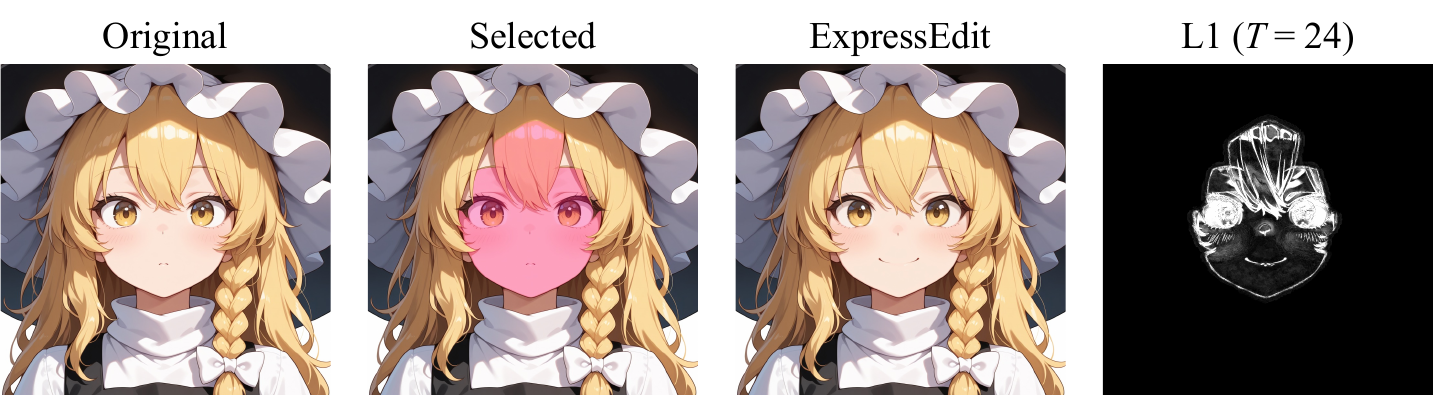} 
      \caption{\name\ introduced strictly no noise outside the edited region. The pink color patch on the face shows the selected and edited region.}
      \label{fig:noise_ours}
  \end{subfigure}
  \begin{subfigure}{1\columnwidth}
      \centering
      \includegraphics[width=1\columnwidth]{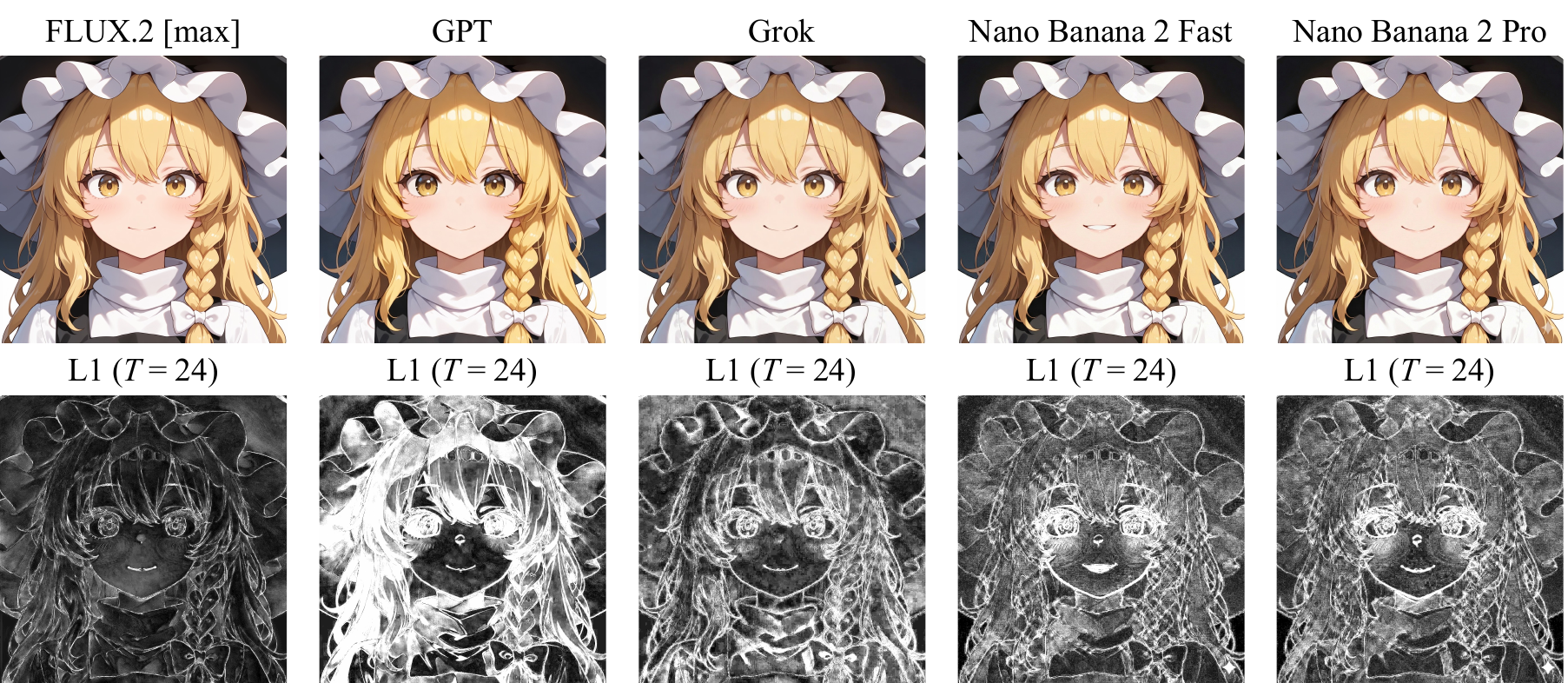} 
      \caption{The 5 baseline models introduced noise globally, sometimes with specific watermark patterns. The L1 distance between the original and each edited image is visualized to highlight the noise patterns. The L1 distance is linearly mapped to grayscale colors between black and white, with a threshold $T = 24$.}
      \label{fig:noise_baseline}
  \end{subfigure}
  \begin{subfigure}{1\columnwidth}
      \centering
      \includegraphics[width=1\columnwidth]{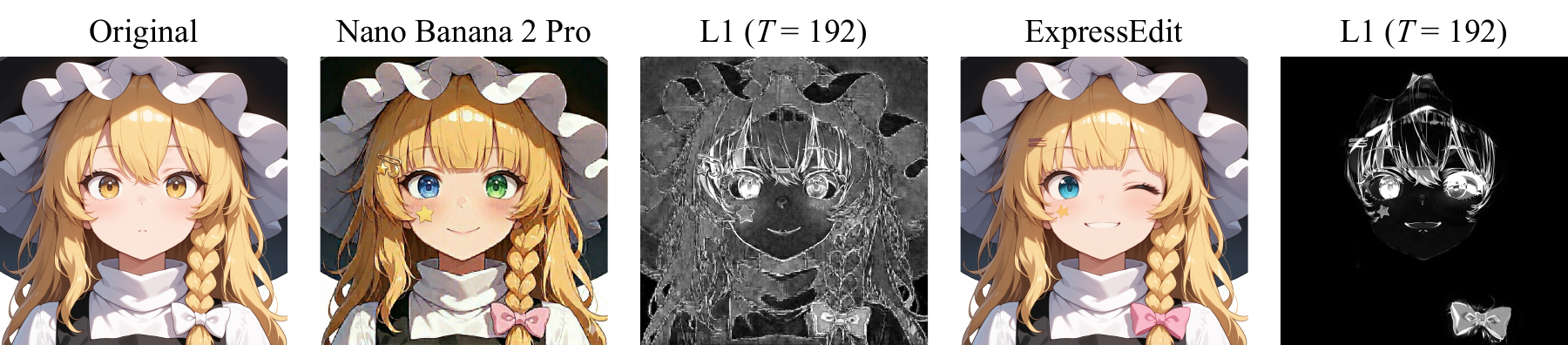} 
      \caption{In 8 steps, the noise from Nano Banana 2 Pro corrupted the image. Nano Banana 2 Pro failed to make the character wink on Step 7 (\Cref{tab:eight_prompts}). }
      \label{fig:eight_steps}
  \end{subfigure}
  \begin{subfigure}{1\columnwidth}
      \centering
      \includegraphics[width=1\columnwidth]{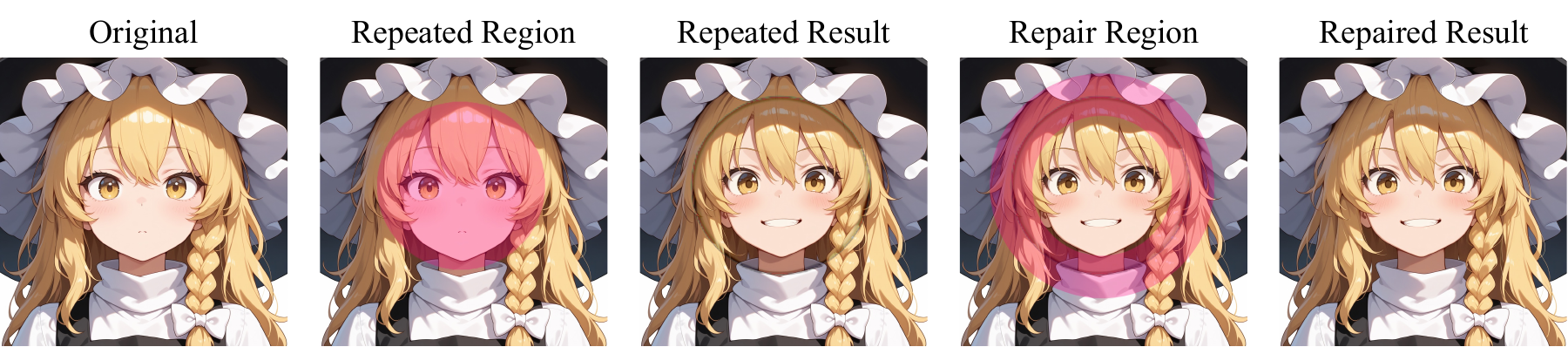} 
      \caption{In 100 steps, \name\ introduces noise only around the selection edge, and the noise is easily repaired in one step.}
      \label{fig:repair}
  \end{subfigure}
  \caption{\textbf{Baseline methods introduce destructive noise into the original image after each editing step, whereas the pixel changes from \name\ are non-destructive, and the minor artifacts are easily repaired.} Please see \Cref{sec:clean-edits} for details.}
  \label{fig:noise}
\end{figure}

Despite their strong prompt-following performance, baseline models introduce heavy noise in the image regions that should not be edited according to the prompt. As an example, when the user wants to edit an expression to smiling, the hair and clothes of the character should not be touched (\Cref{fig:noise_ours}). In \name, the selection of the face is made by clicking on the face and dragging slightly, using \photoshop{Quick Selection}~\cite{adobe2026paintselection}. Even though the selection edges are hard without smoothing operations such as \photoshop{Feather}~\cite{adobe2026refinesoften}, \photoshop{Defringe}~\cite{adobe2026fringepixels}, or \photoshop{Expand}~\cite{adobe2026expandcontract}, \name\ cleanly edits the face without visible artifacts. However, baseline models introduced visible noise all over the image (\Cref{fig:noise_baseline}). The noise might appear less distracting in photo-realistic images, but it is much more easily identifiable in the clean colors of stylized animation images, as there are fewer high-frequency details~\cite{wang2024apisr}.

To highlight the noise patterns, we calculate and visualize the L1 distance in the RGB space (each channel from 0 to 255) between the original and edited images. Pixels with L1 distance between 0 and the threshold value $T$ are mapped linearly to grayscale colors from pure black to pure white, and all pixels with L1 distances larger than $T$ are mapped to pure white. This visualization reveals a much larger color drift from GPT, along with a distinct diagonal noise pattern from the two Nano Banana 2 models.

While one may argue that the noise is negligible to untrained human eyes and thus unimportant, the noise creates a practical challenge for users. Multi-step, iterative editing is an inherent nature of creative workflows~\cite{agarwala2004interactive}. As editing progresses with more steps, the small noise at each step quickly accumulates into corruptions over the whole image (\Cref{fig:eight_steps}). The eight prompts are shown in \Cref{tab:eight_prompts}.

\begin{table*}
  \small
  \caption{\textbf{To iteratively edit one image, \name\ accepts succinct but informative prompts.} For \name, the prompt suffix and prefix (\Cref{fig:pipeline-image}) are kept fixed, and the tags can be updated minimally to reflect any change on the character. If the selection potentially covers more than one facial element, additional descripions can be added to stabilize the results. For example, when editing the bangs to be blunt in Step 5, the selection may touch the eyes, so adding eye color tags helps prevent the eye color from changing. Notably, no description of composition is needed. For each step, the region of interest is indicated by native Photoshop operations, instead of by text. The prompt for \name\ also does not need to strictly follow a set of existing tags, but the tag format empirically leads to better results.}
  \label{tab:eight_prompts}
  \centering
  \begin{tabular}{@{}cll@{}}
    \toprule
    \textbf{Step} & \textbf{Prompt for Baseline Models} & \textbf{Prompt for \name\ (Prefix and Suffix Omitted)} \\
    \midrule
    1 & Change the color of her left eye to green & green eye \\
    2 & Change the color of her right eye to blue & blue eye \\
    3 & Make her smile & green eye, blue eye, smile \\
    4 & Change the color of her bow to pink & pink bow \\
    5 & Make her bangs blunt & green eye, blue eye, blunt bangs \\
    6 & Add a yellow star sticker under her right eye & blue eye, yellow star sticker \\
    7 & Close her left eye to make her wink & wink \\
    8 & Add a hairpin & hairpin \\
    \bottomrule
  \end{tabular}
\end{table*}

\name, however, does not accumulate noise in this case. This is one key benefit of the denoising process in the diffusion model backend. The adoption of an open-source backend also prevents the injection of watermarks~\cite{gowal2025synthid, googledeepmind2026synthid}, which further degrades image quality beyond the user's control. Even in a stress-testing case where \photoshop{selections} strictly overlap over 100 steps, \name\ only accumulates noise around the edge of the selection (\Cref{fig:repair}), and the noise can be easily removed within one step. The user only needs to \photoshop{select} the noisy region, and the \plugin{prompt} is kept fixed.

\begin{figure}
  \centering
  \includegraphics[width=1\columnwidth]{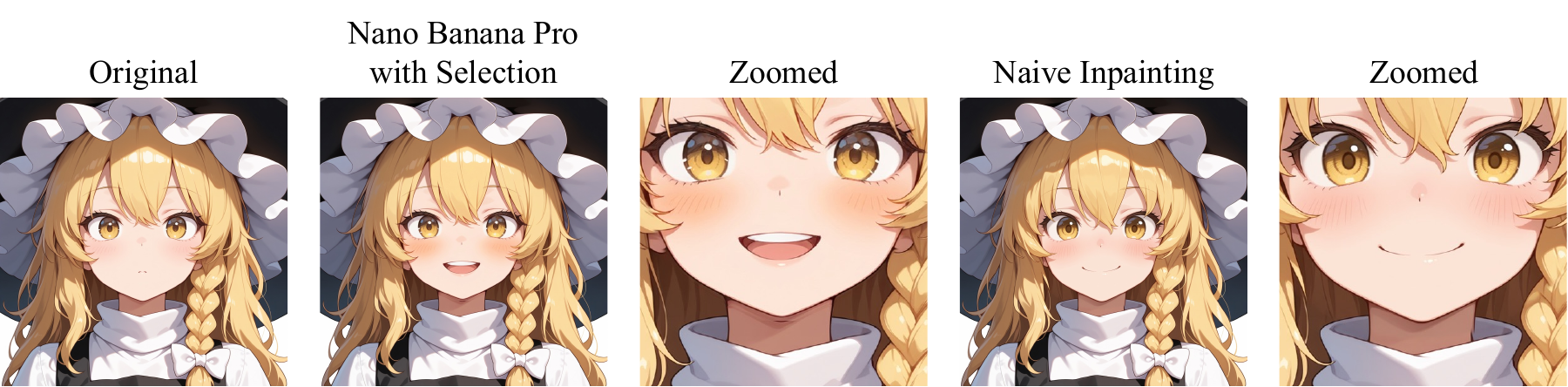}
  \caption{\textbf{The selection version of Nano Banana Pro in Photoshop and the naive inpainting both create visible artifacts around the selected region.} Nano Banana Pro leaves artifacts around earlobes and the chin, making it impossible to contain the destructive noise via selecting. Naive inpainting also leaves artifacts on the right side of the neck and on the braid, necessitating the use of the SPICE backend.}
  \label{fig:local}
\end{figure}

One may also argue that the noise for baseline models could be contained within the selected region if a selection were provided. However, one weakness of the baseline models forbids this operation. As an example, the official integration of Nano Banana Pro in Photoshop supports editing a selected region.\footnote{As of March 2026, Nano Banana 2 Pro has not been integrated into Photoshop. Only the first version of Nano Banana Pro is available~\cite{adobe2026generativefill}.} When the selected region is edited, the edges in the selected region frequently mismatch the original edge, necessitating manual post-processing (\Cref{fig:local}). This is due to the pixel drifting problem commonly observed in recent image editing models~\cite{ye2026agent}. As the naive inpainting method using diffusion models has a similar effect, we uses the SPICE backend with explicit Canny edge control to eliminate this weakness~\cite{tang2025spice}. Notably, when the selection is drawn with full \photoshop{Hardness} using the \photoshop{Selection Brush}~\cite{adobe2024lassotools} as in \Cref{fig:pipeline-image}, no edge artifacts are created by \name\ in the generated result. 

\begin{figure}
  \centering
  \includegraphics[width=1\columnwidth]{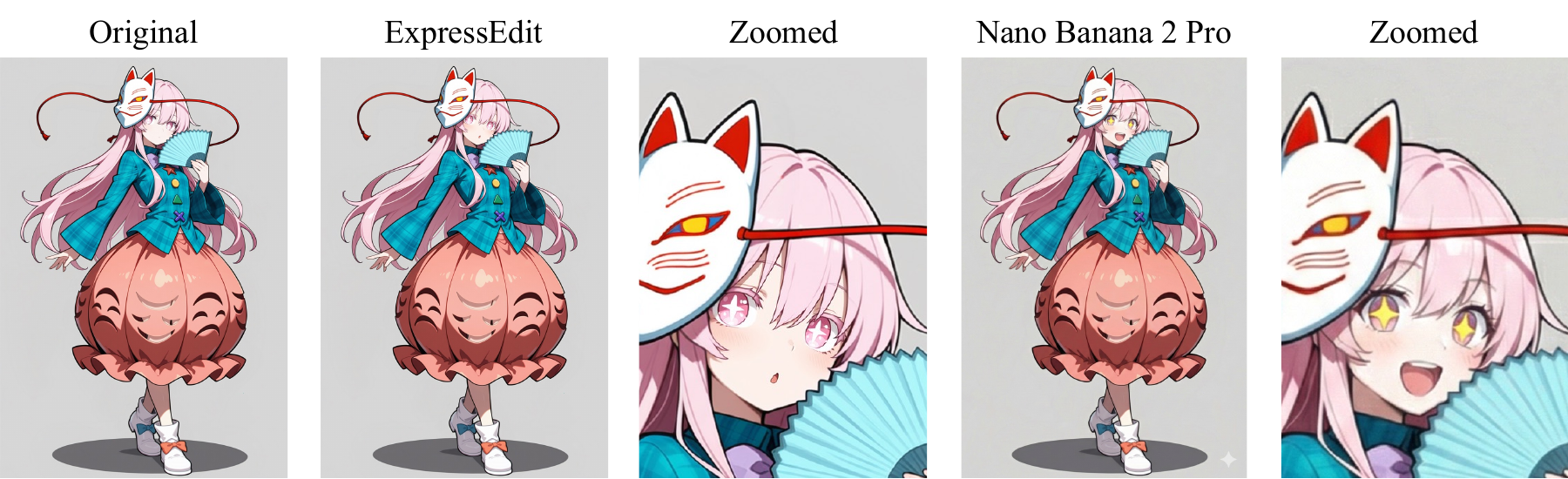}
  \caption{\textbf{\name\ can generate detailed expressions even with challenging, high-resolution inputs, whereas Nano Banana 2 Pro generates lower quality faces, despite its support for 2K image generation.} Nano Banana 2 Pro generates a face with blurry contours, creates visible artifacts around the red string, lowers the saturation, and arbitrarily changes the shape of the face.}
  \label{fig:large}
\end{figure}

\photoshop{Selection} also allows ExpressEdit to operate on high resolution images. \Cref{fig:large} shows an example of editing an 1664$\times$2432 image, with two expression tags ``\texttt{+\_+}'' and ``\texttt{:O}''. The prompt for Nano Banana 2 Pro is ``Make her excited with open mouth, eyes lighting up in excitement, with a yellow four-pointed sparkle in the center.'' For large images, while Nano Banana 2 Pro supports high resolution output, the output was still degraded in various aspects, such as reduced saturation and unwanted deformations.

\subsection{Responsive and Precise Edits}
\label{sec:responsive-and-precise-edits}

\begin{figure}
  \centering
  \includegraphics[width=1\columnwidth]{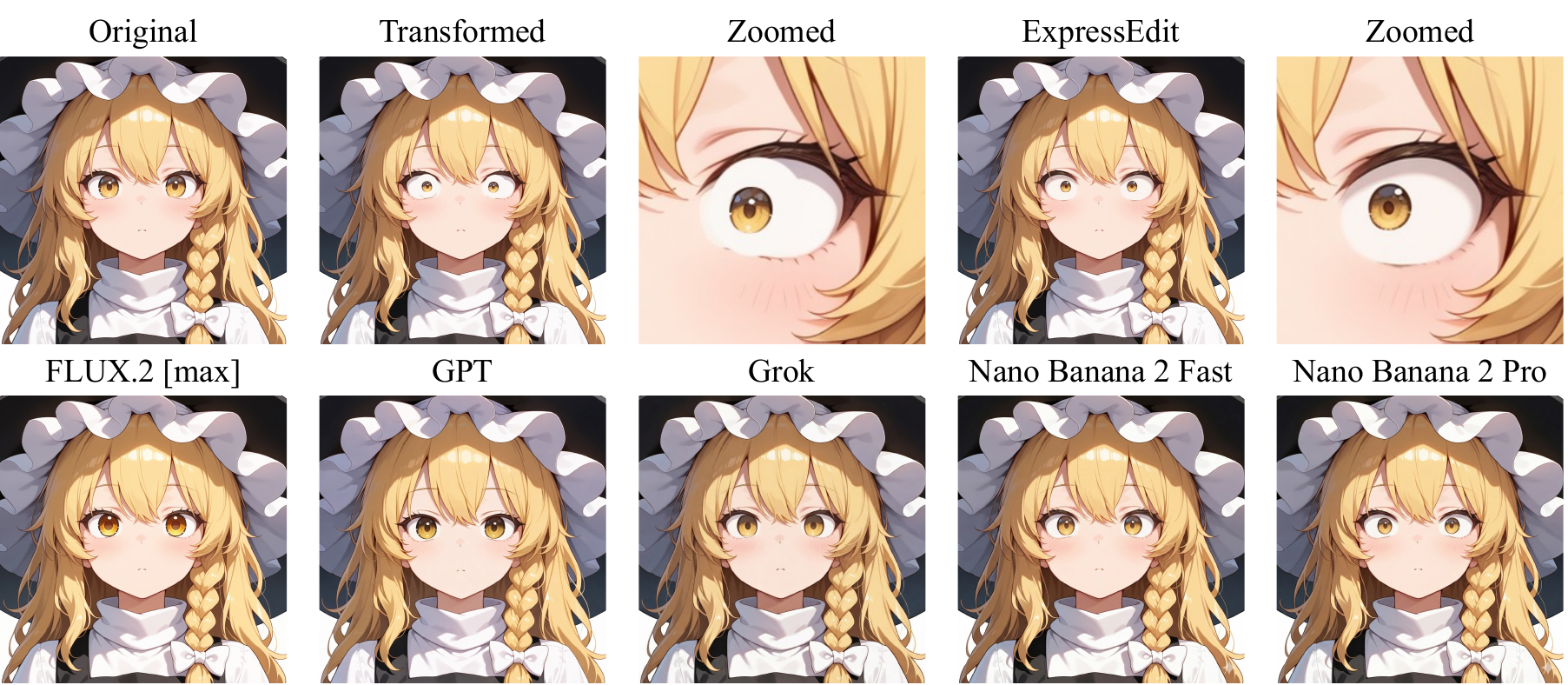}
  \caption{\textbf{\name\ precisely follows the instruction (reducing the diameter of the irises by 50\%) with a simple transformation in Photoshop as a hint to the diffusion model.} To reduce the iris size, the user only needs to select the irises, transform their size, and leave the holes from transformation in a white color. There is no need to manually re-draw the shadows on the eyeball, as \name\ automatically fixes the gap. While given a clear instruction specifying the percentage of size change, all baseline models fail on this task. \name, however, correctly generates the result, even without the numeric number or expression tags in the prompt. Only the prompt prefix for the character and the prompt suffix for the style were used.}
  \label{fig:half_iris}
\end{figure}

In stylized expressions, fine-grained editing of facial elements is sometimes required to precisely convey the extent of an emotion. For example, the size of the iris can be reduced to show surprise or horror~\cite{kanade2000comprehensive}. However, all baseline models fail on this task (\Cref{fig:half_iris}), not responding to the precise numeric description in the prompt (``Reduce the diameter of both irises to 50\% of their current size'').

Assisted by native Photoshop operations, \name\ allow users of all skill levels to easily achieve the desired effect with the following quick steps. To reduce the iris size, the user only needs to \photoshop{Select} the irises, use the \photoshop{Scale}~\cite{adobe2026adjustscale} transformation to change their size, and leave the holes from transformation in a white color. There is no need to manually re-draw the shadows on the eyeball, as \name\ automatically fixes the gap. The user also does not need to specify the numeric details in the prompt, as the RGB-space hint suffices as a hint. In this case, we only used the prompt prefix and suffix, without expression tags at all.

Besides iris size, this sequence of operations can be applied to the size and location of all other facial elements~\cite{kanade2000comprehensive} as well. In this manner, \name\ precisely controls the emotion scale, without dedicated sliders for individual expressions or emotions~\cite{jain2025adaptivesliders}.

\subsection{Quick Synergy with the Liquify Tool}
\label{sec:quick-synergy-with-the-liquify-tool}

As an alternative to \photoshop{Select} and \photoshop{Scale}, directly dragging elements to their desired locations is more intuitive for editing an image. This intuitive editing operation corresponds to the \photoshop{Liquify} tool in Photoshop, and has motivated the training of many AI-based image editing models~\cite{mou2024dragondiffusion, liu2024drag, ling2024freedrag}. 

\begin{figure}
  \centering
  \includegraphics[width=1\columnwidth]{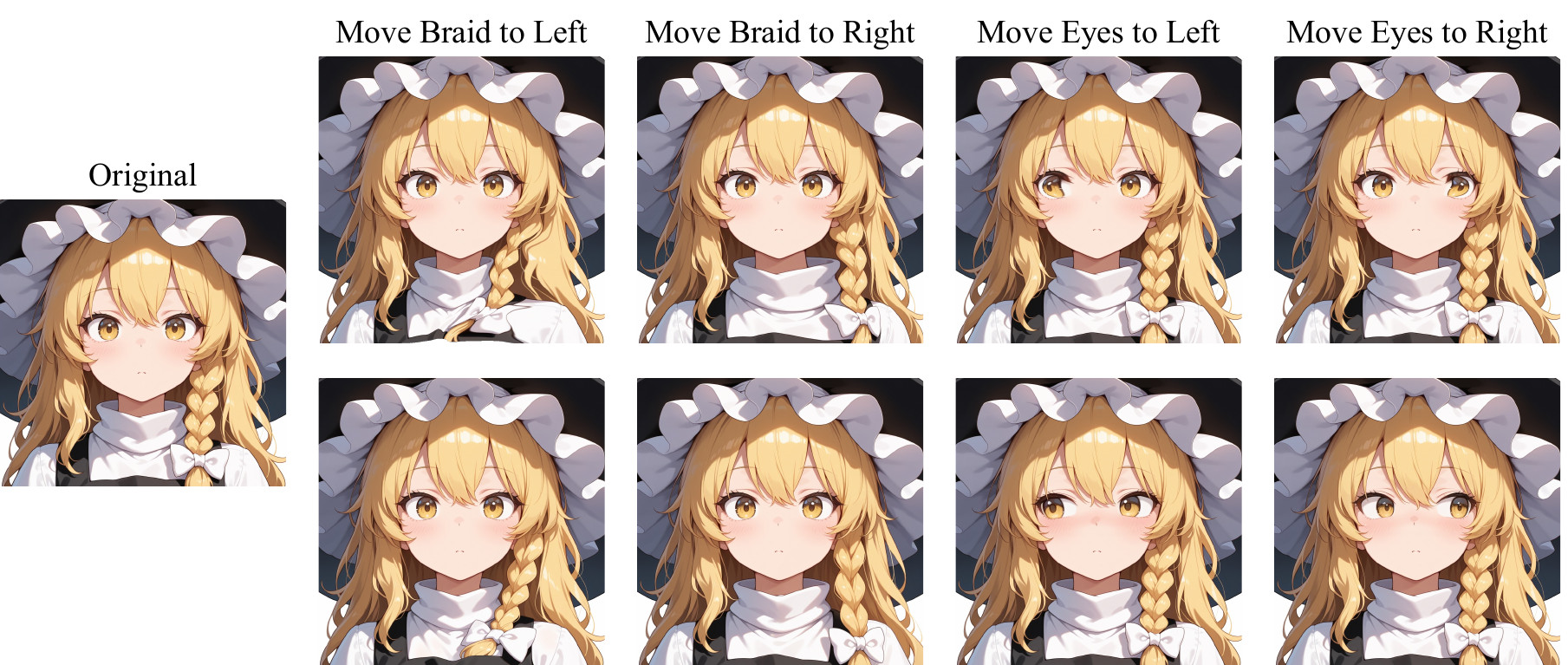}
  \caption{\textbf{\name\ conveniently fixes artifacts from manual editing.} The first row shows the results from the Liquify tool in Photoshop, and the second row shows the fixed results. Liquify supports high editing flexibility at the notorious cost of long manual editing time. A casual use of the tool introduces heavy deformations on the white bow or on the iris, but the deformation can be quickly fixed.}
  \label{fig:liquify}
\end{figure}

However, only relying on \photoshop{Liquify} for editing requires significant manual effort. A quick use of \photoshop{Liquify} will leave heavy deformation artifacts on the images, such as a dent on the iris (\Cref{fig:pipeline-image}). Moreover, AI-based image editing models trained for dragging are not adaptable to diverse editing tasks. \name\ overcomes these challenges by using a backend that is robust enough to handle general-purpose editing and artifact repairing at the same time. As shown in \Cref{fig:liquify}, extreme distortions caused by casual \photoshop{Liquify} can be repaired into natural results. In fact, \name\ even benefits from \photoshop{Liquify} as there is no need to specify the left and right directions in the prompt, which are hard for multi-modal models to identify~\cite{saharia2022photorealistic, huang2023t2i, wolf2025your} due to intrinsic limitations of the underlying CLIP model~\cite{kang2025clip}.

The robustness to distortion artifacts also makes the editing process less reliant on \photoshop{Layers}~\cite{adobe2022createlayers} or dedicated layering models~\cite{yin2025qwen}, when the edited regions overlap with other objects. Nevertheless, \name\ operates on \photoshop{Visible Layers}~\cite{adobe2026samplefrom}, still enabling the editing of only certain layers should the user find it necessary.

\subsection{High Adaptability to Broader Edits}

\begin{figure}
  \centering
  \includegraphics[width=1\columnwidth]{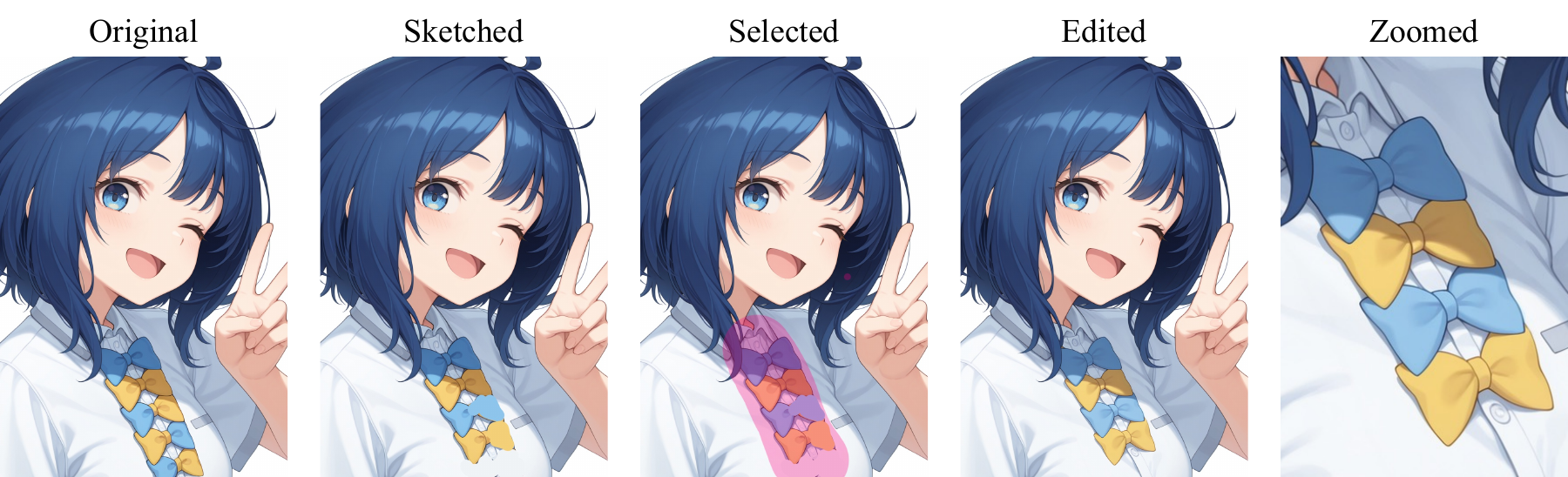}
  \caption{\textbf{Besides expressions, \name\ can fix other details on a character.} The correct design of the character includes 4 bows with interleaving blue and yellow colors~\cite{makeine2026character}. Hinted by simple sketches, \name\ fixes the design in a few seconds.}
  \label{fig:adaptability}
\end{figure}

While \name\ excels at editing expressions, it can also edit artifacts specific to AI-generated images. For example, complex designs of characters are usually not generated correctly, with artifacts such as incorrect number of accessories or scrambled colors (\Cref{fig:adaptability}). These deviations significantly interfere with the conveyed character identity~\cite{cheng2024evaluating, zell2019perception, singhania2025beyond, shakirov2024impact, hester2023dress}. With \name, a user can fix this error without advanced digital painting knowledge. By simply sketching the desired pattern on the image using the \photoshop{Color Picker}~\cite{adobe2022choosecolors} and the \photoshop{Hard Round Brush}~\cite{adobe2022setupbrushes}, the user can instruct \name\ to fix the artifacts, strictly maintaining character consistency. Alternatively, the color of the bow-tie can be changed using \photoshop{Adjust Hue/Saturation}~\cite{adobe2024adjustingcolor}, which leads to similar results.

\subsection{Fast Inference with Speed-Up LoRAs}
\label{sec:fast-infererence-with-speed-up-loras}

\begin{figure}
  \centering
  \includegraphics[width=1\columnwidth]{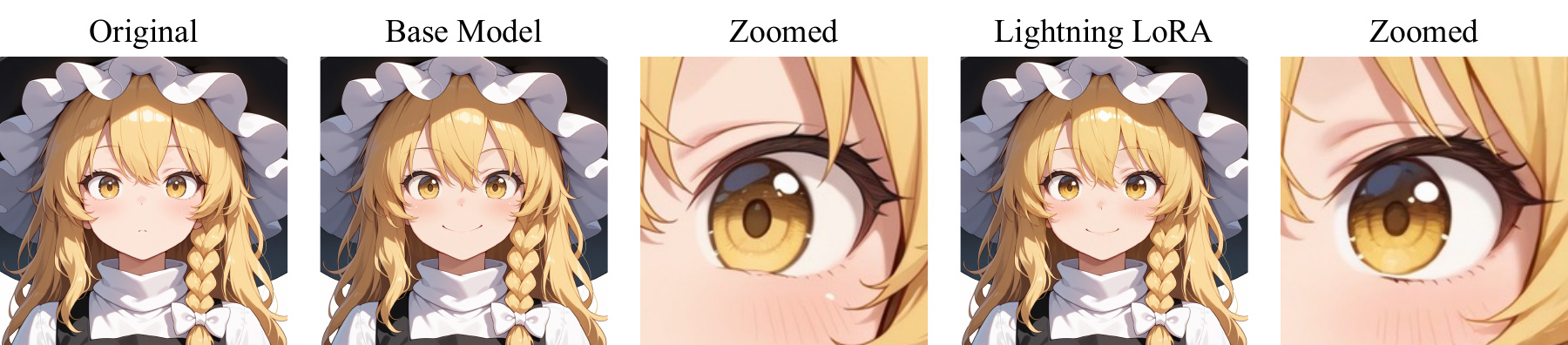}
  \caption{\textbf{Using a speed-up (lightning) LoRA dramatically cuts the latency to 2.18 seconds, with negligible impact on the level of details.} \name\ with a lighting LoRA achieves 46\% lower latency, at the cost of small artifacts on eyelashes.}
  \label{fig:lightning}
\end{figure}

Users with limited compute often use speed-up LoRAs~\cite{ren2024hyper, luo2023latent} to reduce the number of inference steps for faster generation. To support this need from the user, \name\ works seamlessly with speed-up LoRAs, requiring the following three steps from the user: putting the LoRA in the LoRA folder in the backend, adding the trigger words in the prompt, and adjusting the steps and CFG scale in the settings panel. With a speedup LoRA that reduces sampling steps from 30 to 8~\cite{lin2024sdxl, civitai2024sdxllightning}, \name\ reduces API latency by 46\% from 4.06 seconds to 2.18 seconds, faster than all baseline models (\Cref{tab:latency}). While running a total of 30 steps achieves higher visual quality and better details (\Cref{fig:lightning}), speed-up LoRAs can be used for faster prototyping.

\begin{table}
\small
\caption{\textbf{Besides its high quality, \name\ also achieves lowest inference latency using a single consumer-grade GPU.} The latencies are evaluated on an 1024$\times$1024 image. This table shows mean $\pm$ standard deviation over 10 editing runs. The latency was measured from clicking Generate to the image appearing. Due to the open source nature of \name, we were able to test it under a stable environment without the interference of other resource-intensive software, resulting in small standard deviations. An additional overhead of 1 to 2 seconds should be expected if the user uses slower devices, or if the user uses other software for painting assistance at the same time. Still, \name\ is the fastest, and it generates the cleanest results (\Cref{sec:clean-edits}) without any API cost.}
  \label{tab:latency}
  \centering
  \begin{tabular}{@{}lc@{}}
    \toprule
    \textbf{Method} & \textbf{Latency (s)} \\
    \midrule
    FLUX.2 [max] & 49.94 $\pm$ 13.39 \\
    GPT & 46.01 $\pm$ 11.74\\
    Grok & 7.11 $\pm$ 0.50 \\
    Nano Banana 2 & \\
    \quad Fast (without Reasoning) & 23.18 $\pm$ 3.92 \\
    \quad Pro (with Reasoning) & 41.92 $\pm$ 22.08\\
    \midrule
    \name\  & \\
    \quad 30 Sampling Steps & 4.06 $\pm$ 0.02 \\
    \quad 8 Sampling Steps (with Speed-Up LoRA) & \textbf{2.18 $\pm$ 0.02} \\
    \bottomrule
  \end{tabular}
\end{table}

Besides speed-up LoRAs, \name\ also supports character, expression, and style LoRAs. Due to the page limit, we only demonstrate the result from one character LoRA~\cite{littlejelly2025yanamianna} in \Cref{fig:adaptability}.

\section{Conclusion}

In this paper, we present \name, an open-source Photoshop plugin that efficiently edits stylized expressions. Assisted by a large database of expression tags, \name\ generates clean images without noise or watermarks. Moreover, the seamless integration into Photoshop allows the user to take full advantage of the powerful native operations, even cutting the time spent on traditionally time-consuming operations. We open source the full dataset and code to facilitate future research and artistic exploration.

{
    \small
    \bibliographystyle{ieeenat_fullname}
    \bibliography{main}
}


\end{document}